\documentclass[sigconf]{acmart}
\AtBeginDocument{%
  }

\renewcommand\footnotetextcopyrightpermission[1]{}
\usepackage{amsmath} 
\usepackage{amsfonts} 
\usepackage{booktabs}
\usepackage{multirow}
\usepackage{makecell}
\usepackage{xcolor}
\usepackage{bbding}
\usepackage{algorithm}
\usepackage{algorithmic}
\usepackage{cleveref}
\usepackage{tablefootnote}
\usepackage{colortbl}
\usepackage{pifont}
\begin{document}

\title{Efficiency Meets Fidelity: A Novel Quantization Framework for Stable Diffusion}

\author{Shuaiting Li}
\authornote{Both authors contributed equally to this research.}
\email{list@zju.edu.cn}
\affiliation{%
  \institution{Zhejiang University}
  \city{Hangzhou}
  \state{Zhejiang}
  \country{China}
}

\author{Juncan Deng}
\authornotemark[1]
\email{dengjuncan@zju.edu.cn}
\affiliation{%
  \institution{Zhejiang University}
  \city{Hangzhou}
  \state{Zhejiang}
  \country{China}
}

\author{Zeyu Wang}
\email{wangzeyu2020@zju.edu.cn}
\affiliation{%
  \institution{Zhejiang University}
  \city{Hangzhou}
  \state{Zhejiang}
  \country{China}
}

\author{Kedong Xu}
\email{xukedong@vivo.comn}
\affiliation{%
  \institution{vivo Mobile Communication Co., Ltd}
  \city{Hangzhou}
  \state{Zhejiang}
  \country{China}
}

\author{Rongtao Deng}
\email{dengrongtao@vivo.comn}
\affiliation{%
  \institution{vivo Mobile Communication Co., Ltd}
  \city{Hangzhou}
  \state{Zhejiang}
  \country{China}
}

\author{Hong Gu}
\email{guhong@vivo.comn}
\affiliation{%
  \institution{vivo Mobile Communication Co., Ltd}
  \city{Hangzhou}
  \state{Zhejiang}
  \country{China}
}

\author{Haibin Shen}
\email{shen_hb@zju.edu.cn}
\affiliation{%
  \institution{Zhejiang University}
  \city{Hangzhou}
  \state{Zhejiang}
  \country{China}
}

\author{Kejie Huang}
\authornote{Corresponding author.}
\email{huangkejie@zju.edu.cn}
\affiliation{%
  \institution{Zhejiang University}
  \city{Hangzhou}
  \state{Zhejiang}
  \country{China}
}

\renewcommand{\shortauthors}{Shuaiting Li et al.}

\begin{abstract}
  Text-to-image generation of Stable Diffusion models has achieved notable success due to its remarkable generation ability. However, the repetitive denoising process is computationally intensive during inference, which renders Diffusion models less suitable for real-world applications that require low latency and scalability. Recent studies have employed post-training quantization (PTQ) and quantization-aware training (QAT) methods to compress Diffusion models. Nevertheless, prior research has often neglected to examine the consistency between results generated by quantized models and those from floating-point models. This consistency is crucial in fields such as content creation, design, and edge deployment, as it can significantly enhance both efficiency and system stability for practitioners. To ensure that quantized models generate high-quality and consistent images,  we propose an efficient quantization framework for Stable Diffusion models. Our approach features a Serial-to-Parallel calibration pipeline that addresses the consistency of both the calibration and inference processes, as well as ensuring training stability. Based on this pipeline, we further introduce a mix-precision quantization strategy, multi-timestep activation quantization, and time information precalculation techniques to ensure high-fidelity generation in comparison to floating-point models.
  
  Through extensive experiments with Stable Diffusion v1-4, v2-1, and XL 1.0, we have demonstrated that our method outperforms the current state-of-the-art techniques when tested on prompts from the COCO validation dataset and the Stable-Diffusion-Prompts dataset. Under W4A8 quantization settings, our approach enhances both distribution similarity and visual similarity by 45\%$\sim$60\%. 
\end{abstract}



\begin{abstract}
  Text-to-image generation via Stable Diffusion models (SDM) have demonstrated remarkable capabilities. However, their computational intensity, particularly in the iterative denoising process, hinders real-time deployment in latency-sensitive applications. While Recent studies have explored post-training quantization (PTQ) and quantization-aware training (QAT) methods to compress Diffusion models, existing methods often overlook the consistency between results generated by quantized models and those from floating-point models. This consistency is paramount for professional applications where both efficiency and output reliability are essential. To ensure that quantized SDM generates high-quality and consistent images,  we propose an efficient quantization framework for SDM. Our framework introduces a Serial-to-Parallel pipeline that simultaneously maintains training-inference consistency and ensures optimization stability. Building upon this foundation, we further develop several techniques including multi-timestep activation quantization, time information precalculation, inter-layer distillation, and selective freezing,  to achieve high-fidelity generation in comparison to floating-point models while maintaining quantization efficiency.
  
 Through comprehensive evaluation across multiple Stable Diffusion variants (v1-4, v2-1, XL 1.0, and v3), our method demonstrates superior performance over state-of-the-art approaches with shorter training times. Under W4A8 quantization settings, we achieve significant improvements in both distribution similarity and visual fidelity, while preserving a high image quality.
\end{abstract}
\maketitle

\section{Introduction}

Diffusion models have yielded remarkable achievements and demonstrated exceptional performance across various generative tasks,~\cite{chen2024disenboothidentitypreservingdisentangledtuning, ho2020denoising, rombach2022high, ruiz2023dreambooth, song2022denoisingdiffusionimplicitmodels, song2021scorebasedgenerativemodelingstochastic}, particularly in the realm of text-to-image generation~\cite{chen2024disenboothidentitypreservingdisentangledtuning, rombach2022high, ruiz2023dreambooth}. Nonetheless, these models often entail significant computational expenses, primarily due to two factors.  Firstly, within a Diffusion model, a UNet~\cite{dhariwal2021diffusion, ronneberger2015u} carries out a time-consuming iterative sampling process to progressively denoise a random latent variable. Secondly, the pursuit of superior image quality and higher resolutions has resulted in larger model sizes, necessitating extensive time and memory resources. These challenges render Diffusion models (e.g., Stable Diffusion~\cite{rombach2022high} and Stable Diffusion XL~\cite{podell2023sdxlimprovinglatentdiffusion}) computationally demanding and difficult to deploy in real-world applications requiring low latency and scalability. 

Recently, many researchers have investigated quantization strategies for compressing Diffusion models~\cite{he2024ptqd, li2023q, shang2023post, wang2024accurateposttrainingquantizationdiffusion, tang2024posttrainingquantizationtexttoimagediffusion, he2023efficientdm, sui2024bitsfusion}, predominantly utilizing Post-Training Quantization (PTQ)~\cite{li2021brecqpushinglimitposttraining, nagel2020up}. PTQ does not require a whole dataset and heavy retraining, making it more appealing than Quantization-Aware Training (QAT)~\cite{nagel2022overcoming} for large models. However, PTQ methods on diffusion models experience substantial performance degradation at 4 bits and below. And their calibration processes remain computationally intensive building upon techniques like  AdaRound~\cite{nagel2020up} and BRECQ~\cite{li2021brecqpushinglimitposttraining}. The calibration process of large text-to-image models such as Stable Diffusion XL 1.0 still requires 1 day due to the necessary block-wise training procedures.

Meanwhile, most existing research focuses on optimizing quantized models for high-quality image generation, often neglecting the consistency between quantized and floating-point outputs. In content creation and design, such consistency is critical—quantized models must closely match the style and content of their floating-point counterparts. Otherwise, users face difficulties in predicting and controlling results, requiring extensive prompt tuning and adjustments, which hampers productivity and creative workflow.
Moreover, discrepancies in generated styles can degrade the performance of downstream tasks~\cite{zhang2023adding, wu2024seesr} and undermine system reliability.

To overcome these limitations, we propose a novel Stable Diffusion quantization framework that is specifically designed to achieve high fidelity and efficiency. Through systematic analysis of existing pipelines for the joint optimization of quantized diffusion models, we identify two critical barriers that hinder high-fidelity quantization: \ding{182} Timestep-sequential fine-tuning leads to periodic oscillations in weight gradients, which in turn results in substantial instability in quantized weights.
\ding{183} Distributional discrepancy between noise-scheduled training latents and inference-phase latents leads to activation misalignment.
Addressing the revealed challenges, we propose our Serial-to-Parallel pipeline, which maintains training-inference consistency and ensures optimization stability. To further enhance fidelity, several techniques are integrated into the pipeline, including the preservation of temporal information, the utilization of multiple time-step activation quantizers, inter-layer distillation on sensitive layers, and a selective weight freezing strategy on these layers.

The generated results are evaluated in terms of similarity, quality, and text-image alignment. In comparison to previous PTQ methods, our framework demonstrates superior generation consistency in shorter training times across UNet-based and MMDiT-based Stable Diffusion models.

Our contributions are summarized as follows:
\begin{itemize}
\item We propose a Serial-to-Parallel pipeline that maintains training-inference consistency and ensures optimization stability.
\item We integrate several techniques into the pipeline for higher generation fidelity.
\item We validate the fidelity and consistency improvements over sota methods.
\end{itemize}
 
\section{Related Work}
\subsection{Diffusion Model Acceleration}
While Stable Diffusion models can generate high-quality samples, their slow generation speeds pose a significant challenge for large-scale applications. To tackle this problem, significant efforts have focused on improving the efficiency of the sampling process, which can be categorized into two methods. 

The first method involves designing advanced samplers for pre-trained models, such as analytical trajectory estimation~\cite{bao2022analyticdpmanalyticestimateoptimal, bao2022estimatingoptimalcovarianceimperfect}, implicit sampler~\cite{song2022denoisingdiffusionimplicitmodels, kong2021fastsamplingdiffusionprobabilistic, zhang2023gddimgeneralizeddenoisingdiffusion, watson2021learningefficientlysamplediffusion}, stochastic differential equations~\cite{song2021scorebasedgenerativemodelingstochastic, jolicoeurmartineau2021gottafastgeneratingdata, kim2022denoisingmcmcacceleratingdiffusionbased} and ordinary differential equations~\cite{lu2022dpm, liu2022pseudonumericalmethodsdiffusion, zhang2023fastsamplingdiffusionmodels}. Although these methods can reduce the number of sampling iterations required, the significant parameter count and computational demands of Stable Diffusion models limit their application on edge devices.

The second method involves retraining the model, such as diffusion scheme optimization~\cite{chung2022come, franzese2023much, zheng2023truncateddiffusionprobabilisticmodels, lyu2022acceleratingdiffusionmodelsearly}, knowledge distillation~\cite{luhman2021knowledgedistillationiterativegenerative, salimans2022progressivedistillationfastsampling},  sample trajectory optimization~\cite{watson2021learningefficientlysamplediffusion, lam2022bddmbilateraldenoisingdiffusion}, and noise scale adjustment~\cite{kingma2021variational, nichol2021improved}. Though these techniques effectively speed up the sampling process, re-training a Diffusion model is computationally intensive, especially for resource-constrained devices.

\subsection{Diffusion Model Quantization}
Quantization is a widely used technique that aids in reducing memory usage and speeding up computation. It is generally categorized into two types: QAT~\cite{gong2019differentiable, louizos2018relaxedquantizationdiscretizedneural, jacob2018quantization, zhuang2018towards, zhang2023root} and PTQ~\cite{li2021brecqpushinglimitposttraining, nagel2020up, hubara2020improvingposttrainingneural, wei2023qdroprandomlydroppingquantization, lin2023fqvitposttrainingquantizationfully}. 
EfficientDM~\cite{he2023efficientdm} is representative of QAT work, it proposes a data-free distillation framework and applies a quantization-aware variant of the low-rank adapter. While training on the whole dataset is time-consuming and computationally heavy, recent studies focus on PTQ for Diffusion models, which only necessitates a small amount of unlabeled data for calibration. PTQ4DM~\cite{shang2023post} and Q-diffusion~\cite{li2023q} focus on sampling the noise of the floating-point model across different timesteps, Q-diffusion further propose to split the activation of shortcut layers. PTQD~\cite{he2024ptqd} disentangle the quantization noise into its correlated and residual uncorrelated parts and correct them individually. PCR~\cite{tang2024posttrainingquantizationtexttoimagediffusion} progressively calibrates the activation quantizer considering the accumulated quantization error across timesteps and selectively relaxing the bit-width for several of those timesteps.
However, these PTQ methods seldom consider the consistency of the generated output, and many of them are not designed for new, large pre-trained text-to-image models, such as Stable Diffusion.

\section{Preliminaries}
\subsection{Diffusion Models}
Diffusion models~\cite{ho2020denoising, song2022denoisingdiffusionimplicitmodels} gradually add Gaussian noise with a variance schedule $\beta_1, \ldots, \beta_T \in$ $(0,1)$ to real image $x_0 \sim q(x)$ for $T$ times as sampling process, resulting in a sequence of noisy samples $x_1, \ldots, x_T$. In DDPMs~\cite{ho2020denoising}, the sampling process is a Markov chain, which can be formulated as:
\begin{equation}
\begin{aligned}
q\left(x_{1: T} \mid x_0\right)&=\prod_{t=1}^T q\left(x_t \mid x_{t-1}\right), q\left(x_t \mid x_{t-1}\right) \\
&=\mathcal{N}\left(x_t ; \sqrt{\alpha_t} x_{t-1}, \beta_t \mathbf{I}\right)
\label{eq1}
\end{aligned}
\end{equation}
where $\beta_t=1-\alpha_t$. Conversely, the denoising process removes noise from a sample from Gaussian noise $x_T \sim$ $\mathcal{N}(\mathbf{0}, \mathbf{I})$ to gradually generate high-fidelity images. However, due to the unavailability of the true reverse conditional distribution $q\left(x_{t-1} \mid x_t\right)$, Diffusion models approximate it via variational inference by learning a Gaussian distribution $p_\theta\left(x_{t-1} \mid x_t\right)=\mathcal{N}\left(x_{t-1} ; \boldsymbol{\mu}_\theta\left(x_t, t\right), \boldsymbol{\Sigma}_\theta\left(x_t, t\right)\right)$, the $\boldsymbol{\mu}_\theta$ can be derived by reparameterization trick as follows:
\begin{equation}
\boldsymbol{\mu}_\theta\left(x_t, t\right)=\frac{1}{\sqrt{\alpha_t}}\left(x_t-\frac{\beta_t}{\sqrt{1-\bar{\alpha}_t}} \boldsymbol{\epsilon}_\theta\left(x_t, t\right)\right)
\end{equation}
where $\bar{\alpha}_t=\prod_{i=1}^t \alpha_i$ and $\boldsymbol{\epsilon}_\theta(\cdot)$ is a trainable model to predict noise. The variance $\Sigma_\theta\left(\mathrm{x}_t, t\right)$ can be either learned~\cite{nichol2021improved} or fixed to a constant schedule~\cite{ho2020denoising} $\sigma_t$. When it uses a constant schedule, $x_{t-1}$ can be expressed as:
\begin{equation}
x_{t-1}=\frac{1}{\sqrt{\alpha_t}}\left(x_t-\frac{\beta_t}{\sqrt{1-\bar{\alpha}_t}} \boldsymbol{\epsilon}_\theta\left(x_t, t\right)\right)+\sigma_t \mathbf{z}
\end{equation}
where $\mathbf{z} \sim \mathcal{N}(\mathbf{0}, \mathbf{I})$.

The formulas outlined in our research are based on the DDPM framework but can be easily adjusted for other accelerated sampling techniques such as DDIM~\cite{song2022denoisingdiffusionimplicitmodels}, PNDM~\cite{liu2022pseudonumericalmethodsdiffusion}, and Euler~\cite{karras2022elucidating}.

In this work, we adopt Denoising Diffusion Block (DDB) as the unified term for the backbone denoising networks $\boldsymbol{\epsilon}_\theta(\cdot)$. Specifically, for SD1.4, SD2.1, and SDXL, the DDB is implemented as a U-Net architecture.
For SD3, the DDB follows the MM-DiT (Multi-Modal Diffusion in Transformer) design.

\subsection{Model Quantization}
Quantization~\cite{nagel2021whitepaperneuralnetwork} is a key technique in model compression. This method compresses neural networks by reducing the number of bits used for model weights and activations. The quantization process can be formulated as:

\begin{equation}
w_{q}=\mathrm{clip}\left(\mathrm{round}\left({\frac{w}{s}}\right)+z,q_{\mathrm{min}},q_{\mathrm{max}}\right)
\end{equation}

where $s$ is the scaling factor, $z$ is the zero-point, and $q_{\mathrm{min}}$ and $q_{\mathrm{max}}$ are the minimum and maximum quantization values, respectively. Reversely, the dequantization process is formulated as:
\begin{equation}
    \hat{w} = (w_{q}-z)\times s
\end{equation}

\section{Methodology}
We begin by analyzing the advantages and disadvantages of previous training pipelines for large pre-trained text-to-image diffusion models. Then we introduce our Serial-to-Parallel training pipeline which combines their advantages. Subsequently, several techniques are integrated into the pipeline. Multi-timestep activation quantizer is set to separately optimize the parameters associated with each timestep. Additionally, the time feature is precalculated and the accurate projection information is saved for training and inference. Furthermore, feature distillation on sensitive layers is employed and selective freezing is applied to those sensitive layers.

\subsection{Serial-to-Parallel Pipeline}
\begin{figure}[t]
    \centering
    \includegraphics[width=0.99\linewidth]{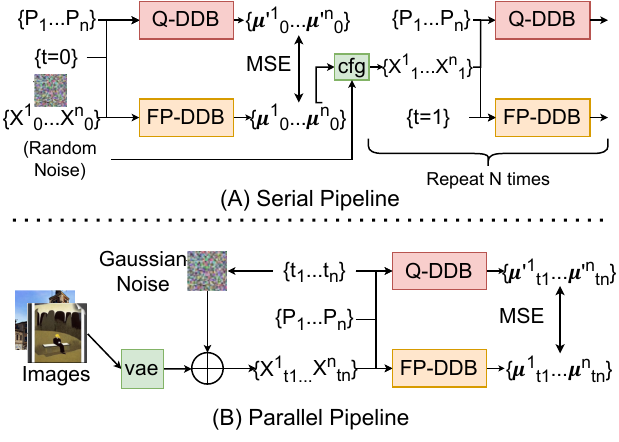}
    \caption{Comparison of (A) 'Serial' pipeline which mimics the inference process of SDM and (B) 'Parallel' pipeline which is more aligned with the pretraining process of SDM. $X_{t}^{n}$ denotes the $n^{th}$ noisy input at timestep $t$. $P$ denotes prompts. $\mu$ denotes the predicted noise. }
    \label{serial_parallel}
\end{figure}
\begin{figure}[t]
    \centering
    \includegraphics[width=0.9\linewidth]{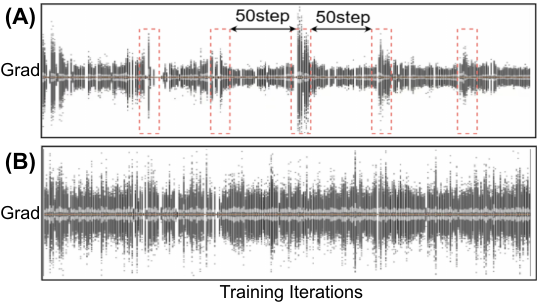}
    \caption{Box plot illustrating the gradient variations in the quantized Stable Diffusion v1-4 model during training. (A) represents the serial pipeline, and (B) represents the parallel pipeline.}
    \label{grad_compare}
\end{figure}

Previous works on jointly optimizing quantized models and distillation-based compression can be roughly divided into two categories: (a) 'Serial'(e.g.~\cite{he2023efficientdm}) and (b) 'Parallel'(e.g.~\cite{kim2023bk,luo2023latentconsistencymodelssynthesizing}), as illustrated in Fig.~\ref{serial_parallel}. The serial pipeline mimics the inference of Stable Diffusion models, and operates in a data-free manner. As shown in Fig.~\ref{serial_parallel}. (A), the initial latent $\{X_{0}^{1}...X_{0}^{n}\}$ are random noises, and only a few prompts are required as input. The latent is denoised step by step based on the prediction of full-precision models, which is sent into quantized DDB. In contrast, the parallel pipeline is more closely aligned with the original Stable Diffusion training process. As shown in Fig.~\ref{serial_parallel}. (B), this approach relies on an image-text pair dataset, where the image is processed through a Variational Autoencoder (VAE) to derive the initial latent. In each iteration, multiple timesteps are randomly sampled, and the latent is then augmented with varying levels of Gaussian noise, as determined by the scheduler. We will present two key observations to reveal their own advantages and disadvantages.

\begin{figure}[t]
    \centering
    \includegraphics[width=1\linewidth]{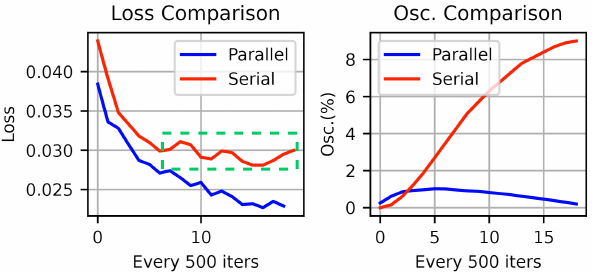}
    \caption{Comparison of loss (left) and weight oscillation (right) between serial and parallel pipeline. Serial pipeline suffers from severe oscillation}
    \label{loss_osc}
\end{figure}

\textbf{\textit{Observation \ding{182}: In serial pipeline, periodic oscillations of gradients leads to severe weight oscillations.}}
As demonstrated in Fig.~\ref{grad_compare}, we have documented the changes in gradients for both serial and parallel pipelines. It can be observed that the gradients remain relatively stable during parallel training, whereas they exhibit periodic oscillations during serial training. Previous research~\cite {reddi2019convergence, wilson2017marginal} have indicated that the Adam optimizer may underperform in the presence of periodic oscillating gradients. 

Building upon Osqat~\cite{nagel2022overcoming}, we systematically analyze weight oscillation patterns across 30 layers, as shown in Fig.~\ref{loss_osc} (right). The serial pipeline demonstrates progressively accumulating weight oscillations, ultimately retaining 9\% unstable parameters - a phenomenon directly correlated with its deteriorating training stability. As evidenced in the green highlighted region of Fig.~\ref{loss_osc} (left), this manifests as persistent loss fluctuations beginning at mid-training that persistently hinder stable convergence. In contrast, the parallel pipeline averages the gradients across multiple time-steps at each iteration and achieves better optimization stability, exhibiting both significantly reduced oscillation rates and smoother loss descent trajectories.

\begin{figure}[t]
    \centering
    \includegraphics[width=1\linewidth]{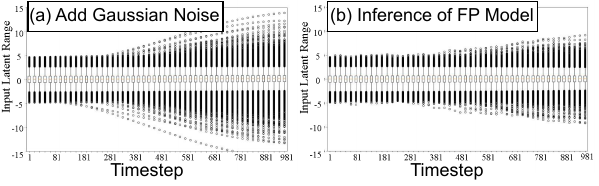}
    \caption{Difference in noisy input range at each timestep with the same initial latent. (a) Adding Gaussian noise based on Eq.~\ref{eq1}. (b) Step-by-step denoising during inference.}
    \label{figure 2}
\end{figure}
\begin{figure*}[t]
    \centering
    \includegraphics[width=0.9\textwidth]{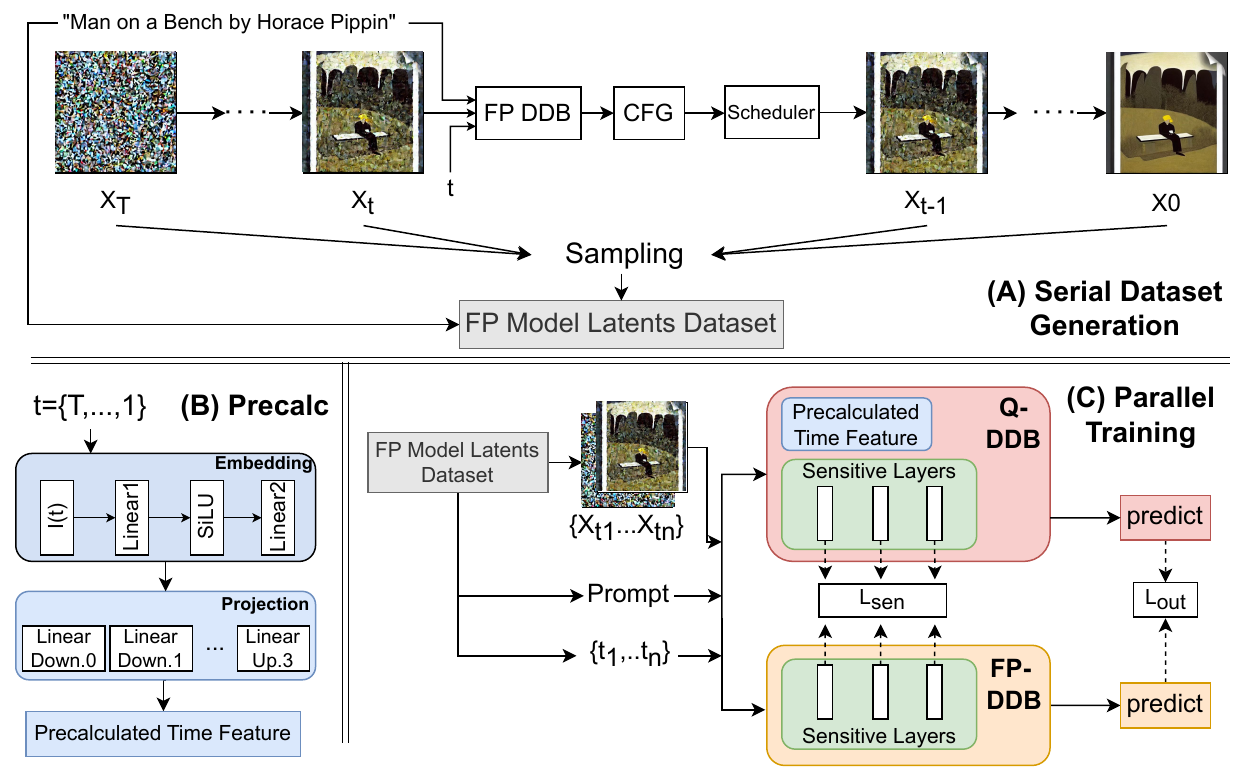}
    \caption{Overview of our quantization framework. (A) Serial dataset generation: During the inference of floating-point models, latent generated from various timesteps for each prompt are randomly sampled. (B) Time information precalculation: The feature map of time projection layers is precalculated for training and inference. (C) Parallel training: At each iteration, latent from various timesteps along with the corresponding prompts are selected from the dataset. The Loss function is calculated between the output and the sensitive layers. Iterative freezing is applied to these sensitive layers for better stability.}
   \label{figure 1}
\end{figure*}
\textbf{\textit{Observation \ding{183}: In parallel pipeline, the input latent range differs between training and inference.}}
During training, noisy inputs are generated by adding Gaussian noise to image-derived latents, whereas inference employs randomly initialized latents undergoing progressive denoising. As shown in Fig.~\ref{figure 2}, these two approaches produce significantly different numerical ranges in the latent representations, that is:
\begin{equation}
\boldsymbol{\mu}_{fp}\left(x_t, t\right)  \neq  x_{t-1} \leftarrow \mathcal{N}\left(x_t ; \sqrt{\alpha_t} x_{t-1}, \beta_t \mathbf{I}\right)
\end{equation}
This mismatch poses challenges for activation quantization, as the quantizer parameters, calibrated on training data, could become suboptimal for inference distributions, potentially amplifying quantization errors. Notably, this issue does not affect the serial pipeline, where training and inference latent ranges remain consistent.
Consequently, it is more beneficial to use the latent sampled from the floating-point model inference as noisy input. 

It can thus be concluded that \textbf{the input latent of the serial pipeline is more appropriate while the training procedure of the parallel pipeline is more reasonable.} To synergize these advantages, we propose our 'Serial-to-Parallel' pipeline that (1) Generates latents serially. As shown in Fig.~\ref{figure 1} (A), the inference is conducted with the floating-point model, whereby the latent is randomly sampled from various timesteps for each prompt. (2) Trains in parallel by utilizing these pre-computed latents. As shown in Fig.~\ref{figure 1} (C), at each iteration, the latent is sampled from different timesteps along with their corresponding prompts from the latent dataset. This strategy renders our framework data-free, relying solely on prompts, while simultaneously enhancing generation consistency and ensuring training stability. The comparison of different pipelines is summarized in Tab.~\ref{pipeline_compare}. 
Our serial-to-parallel pipeline effectively enhances generation consistency while remaining data-free.

\begin{table}[t]
    \centering
    \caption{Comparison of our pipeline and previous pipeline on Stable Diffusion v2.1 with w4a8 quantization. \textit{Osc.}: percentage of oscillating weights at the end of training. Please refer to section5.1 for details of evaluation metrics.}
    \begin{tabular}{cccccc}
    \toprule
    \renewcommand{\arraystretch}{0.9}
    \multirow{2}{*}{Pipeline} & \multirow{2}{*}{Data-free} & \multicolumn{3}{c}{Consistency} & Stability \\
    \cmidrule(lr){3-5}
    \cmidrule(lr){6-6}
    & & FID-FP$\downarrow$ & SSIM$\uparrow$ & LPIPS$\downarrow$ & Osc.(\%)$\downarrow$ \\
    \midrule
    Serial & \checkmark & 13.68 & 0.50 & 0.46 & 9.05 \\
    \midrule
    Parallel & X & 12.92 & 0.53 & 0.42 & 0.33\\
    \midrule
    \makecell[c]{Serial-to\\-Parallel} & \textbf{\checkmark} & \textbf{12.05} & \textbf{0.55} & \textbf{0.40} & \textbf{0.29}\\
    \bottomrule
    \end{tabular}
    \label{pipeline_compare}
\end{table}

\subsection{Components For Higher Fidelity}
Moreover, a variety of techniques are employed to further enhance the fidelity of the generated results.

\textbf{Accurate Activation Quantization.} Previous studies on Diffusion models~\cite{shang2023post, li2023q, so2024temporal, yang2023efficientquantizationstrategieslatent, wang2024questlowbitdiffusionmodel} have shown that the activation distribution at different timesteps varies greatly, posing a challenge for activation quantization.
We adopt different activation quantization parameter sets for different timesteps, which can be expressed as:
\begin{equation}
\mathbf{s}_l = \left\{ s_l^0, s_l^1, \ldots, s_l^{T-1} \right\},\mathbf{z}_l = \left\{ z_l^0, z_l^1, \ldots, z_l^{T-1} \right\}
\end{equation}
where $s_l^t$ and $z_l^t$ are the scaling factor and zero-point of activation quantization parameter for the $l$-th layer at timestep $t$.
The memory consumption of these parameters is negligible and does not influence the inference speed. With regard to the inputs of different time steps within the same batch, our pipeline is capable of efficiently optimizing the activation quantization parameters for these time steps simultaneously.

\textbf{Time Information Precalculation.} 
In a Stable Diffusion model, the time-step $t$ is firstly encoded by time-embedding layers, then passed through time-projection layers in each Bottleneck block. The Time information $e_{p}$ inserted into the UNet is calculated as follows:
\begin{equation}
    e_{t} = emb(t),  ep_{t,i} = proj_{i}(e_{t})
\end{equation}

Consistent with prior work~\cite{huang2024tfmqdmtemporalfeaturemaintenance}, we observe that quantizing time embedding and time projection layers downgrades the quality of the generated images. While TFMQ~\cite{huang2024tfmqdmtemporalfeaturemaintenance} choose to separately optimize these layers, we propose a more fundamental solution. When the inference configuration is determined, the output of the time embedding module $e_t$ is only related to timesteps, and $ep$ is dependent only on $e_t$. Consequently, $ep$ is finite and invariant. Therefore, we remove time-embedding and time-projection from the U-Net and save the $ep$, which is directly input into the model, as demonstrated in Fig.~\ref{figure 1} (B). The memory usage and computational cost of $ep$ are much smaller than the parameters of the time embedding module and the time embedding projection layers.

\textbf{Inter-layer Distillation.}
To enhance generation consistency, we incorporate inter-layer distillation with careful consideration of architectural sensitivities observed in prior work~\cite{wang2024questlowbitdiffusionmodel, sui2024bitsfusion, liu2024taq}. For computational efficiency while preserving performance, we selectively apply distillation to some critical layers in each architecture: (1) in U-Net-based models (SD1.4/2.1/XL), we target shortcut connections and projection layers in feed-forward networks; (2) for MMDiT-based models (SD3), we focus on projection layers in the feed-forward blocks of both flow. 

\begin{figure}[t]
    \centering
    \includegraphics[width=1\linewidth]{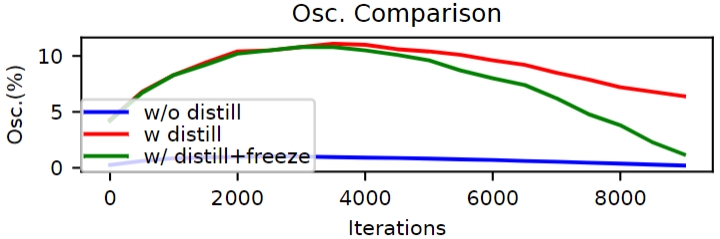}
    \caption{Weight oscillations in selected sensitive layers. The oscillation rates significantly arise after adopting inter-layer distillation, while reduced by iterative freezing.}
    \label{mapping}
\end{figure}

\textbf{Selective Freezing.}
We analyze oscillations~\cite{nagel2022overcoming} of quantized weight in layers chosen for feature distillation and observe a significant increase in instability after inter-layer distillation is applied, as shown in Fig.~\ref{mapping}. To maintain training stability while minimizing computational overhead, we implement a selective iterative freezing mechanism with two key adaptations: (1) reduced freezing frequency (every 500 iterations instead of per-iteration as in~\cite{nagel2022overcoming}), and (2) targeted application to only the mapping layers. This optimized approach introduces negligible runtime overhead (<5\% in SD models) while effectively suppressing distillation-induced oscillations.

\begin{table*}
    \centering
    \caption{Quantitative quality comparisons across different models. Our results outperform other approaches and NF4 baselines. $\dagger$ means reproduced results on our machine. $*$ denotes results in the format of (PCR/Ours). $\downarrow$ means lower is better. \textcolor{red}{$\star$} denotes the key metric. IR is calculated over the first 1000 generated images, while other metrics are calculated over 5000 generated images. }
    \setlength{\tabcolsep}{1.2mm}
    \renewcommand{\arraystretch}{0.95}
    \begin{tabular}{cccccccccccc}
    \toprule
    \multirow{3}{*}{Model} & \multirow{3}{*}{Reso.} & \multirow{3}{*}{Precision} & \multirow{3}{*}{Methods} & \multicolumn{4}{c}{COCO Prompt} &
    \multicolumn{4}{c}{SD prompt} \\
    \cmidrule(lr){5-8}
    \cmidrule(lr){9-12}
    & & & &\multicolumn{2}{c}{Similarity} & Quality & Matching & \multicolumn{2}{c}{Similarity} & Quality & Matching \\
    \cmidrule(lr){5-6}
    \cmidrule(lr){7-7}
    \cmidrule(lr){8-8}
    \cmidrule(lr){9-10}
    \cmidrule(lr){11-11}
    \cmidrule(lr){12-12}
    & & & & \textcolor{red}{$\star$}FID-FP$\downarrow$ & SSIM$\uparrow$ & IR$\uparrow$ &  Clip$\uparrow$ & \textcolor{red}{$\star$}FID-FP$\downarrow$ & SSIM$\uparrow$ & IR $\uparrow$ & Clip$\uparrow$ \\
    \midrule
    \multirow{9}{*}{SD14} & \multirow{9}{*}{512} & FP16 & - & 0 & 1 & 0.16 &  26.48 & 0 & 1 & 0.17 & 28.79*/27.28* \\
    \cmidrule{3-12}
    & & W8A8 & Q-Diffusion & 18.64 & - & -  & 26.15 & 16.22 &- &- & -1.60\\
    & & W8A8 & PTQ4DM & 14.60 &-  &-  & 26,33 & 13.25 &- &- & -1.02\\
    & & W8A8.4 & PCR  & \textbf{8.35} &0.65$\dagger$ & 0.12$\dagger$ & 26.47 & \textbf{9.52} & 0.67$\dagger$ & 0.12$\dagger$ & \textbf{-0.05}\\
    & & \cellcolor{blue!10}W8A8 & \cellcolor{blue!10}Ours & \cellcolor{blue!10}8.57& \cellcolor{blue!10}\textbf{0.69} & \cellcolor{blue!10}\textbf{0.15} & \cellcolor{blue!10}\textbf{26.55} & \cellcolor{blue!10}10.19 & \cellcolor{blue!10}\textbf{0.70} & \cellcolor{blue!10}\textbf{0.14} & \cellcolor{blue!10}-0.14\\
    \cmidrule{3-12}
    & & W4A8 & Q-Diffusion & 20.42 &-  &-  & 26.15& 17.43 & -&- & -1.48\\
    & & W4A8 & PTQ4DM & 17.73 &-  &-  & 26.25& 17.28 & -&- & -1.40\\
    & & W4A8.4 & PCR  & 14.2& 0.47$\dagger$ & -0.05$\dagger$ & 26.48 & 17.9 & 0.50$\dagger$ & -0.28$\dagger$ &-0.74\\
    & & \cellcolor{blue!10}W4A8 & \cellcolor{blue!10}Ours & \cellcolor{blue!10}\textbf{9.46}& \cellcolor{blue!10}\textbf{0.63} &\cellcolor{blue!10}\textbf{0.16} & \cellcolor{blue!10}\textbf{26.49} & \cellcolor{blue!10}\textbf{12.2} & \cellcolor{blue!10}\textbf{0.62} & \cellcolor{blue!10}\textbf{0.11} & \cellcolor{blue!10}\textbf{-0.09}\\
    \midrule
    \multirow{3}{*}{SD21} & \multirow{3}{*}{512} & FP16 & - & 0 & 1 & 0.32 &  26.07 & 0 & 1 & 0.14 & 25.48 \\
    \cmidrule{3-12}
    & & W4A8.4 & PCR & 35.9$\dagger$ & 0.47$\dagger$ & 0.11$\dagger$  &24.14$\dagger$ & 46.3$\dagger$ & 0.49$\dagger$ & -0.44$\dagger$ & 24.18$\dagger$\\
    & & \cellcolor{blue!10}W4A8 & \cellcolor{blue!10}Ours & \cellcolor{blue!10}\textbf{11.60} & \cellcolor{blue!10}\textbf{0.59} & \cellcolor{blue!10}\textbf{0.33} & \cellcolor{blue!10}\textbf{26.08} & \cellcolor{blue!10}\textbf{16.7} & \cellcolor{blue!10}\textbf{0.55} & \cellcolor{blue!10}\textbf{0.11} & \cellcolor{blue!10}\textbf{25.37}\\
    \midrule
    \multirow{5}{*}{SD-XL} & \multirow{5}{*}{768} & FP16 & - & 0 & 1 & 0.49 & 26.51 & 0 & 1 & 0.69 & 29.74*/27.28* \\
    \cmidrule{3-12}
    & & W4A8 & Q-Diffusion & 44.07 & - & - & 15.91 & 24.98 & -& -& -9.81\\
    & & W4A8 & PTQ4DM & 46.45 & - & - & 15.92 & 28.28 & -& -&-9.61\\
    & & W4A8.4 & PCR & 18.27 & 0.48$\dagger$ & -0.35$\dagger$  & 23.85 & 18.25 & 0.52$\dagger$ & -0.15$\dagger$ &-4.03\\
    & & \cellcolor{blue!10}W4A8.4 & \cellcolor{blue!10}Ours & \cellcolor{blue!10}\textbf{13.35} & \cellcolor{blue!10}\textbf{0.61} & \cellcolor{blue!10}\textbf{0.07} & \cellcolor{blue!10}\textbf{24.03} & \cellcolor{blue!10} \textbf{13.08} & \cellcolor{blue!10} \textbf{0.65} & \cellcolor{blue!10} \textbf{0.27} & \cellcolor{blue!10} \textbf{-3.08}\\
    \midrule
    \multirow{12}{*}{SD3} & \multirow{5}{*}{512} & FP16 & - & 0 & 1 & 0.88  & 26.25 & 0 & 1 & 0.91 & 25.86 \\
    \cmidrule{3-12}
    & & \cellcolor{blue!10}W8A8 & \cellcolor{blue!10}Ours& \cellcolor{blue!10}8.99 & \cellcolor{blue!10}0.56 & \cellcolor{blue!10}0.89 &  \cellcolor{blue!10}26.36 & \cellcolor{blue!10}9.02 & \cellcolor{blue!10}0.55 & \cellcolor{blue!10}0.92 & \cellcolor{blue!10}25.89\\
    \cmidrule{3-12}
    & & W4A16 & NF4(g=128) & 25.64 & 0.46 & 0.50 & 25.87 &32.22 &0.42 & 0.69 & 25.70\\
    & & W4A16 & NF4(g=64) & 13.08 & 0.48& 0.72  & 26.12& 16.37 & 0.44& 0.85 & 25.91\\
    & & \cellcolor{blue!10}W4A8 & \cellcolor{blue!10}Ours& \cellcolor{blue!10}\textbf{9.77}& \cellcolor{blue!10}\textbf{0.52} & \cellcolor{blue!10}\textbf{0.90}  & \cellcolor{blue!10}\textbf{26.35} & \cellcolor{blue!10}\textbf{11.03} & \cellcolor{blue!10}\textbf{0.47} & \cellcolor{blue!10}\textbf{0.95} & \cellcolor{blue!10}\textbf{26.09}\\
    \cmidrule{2-12}
    & \multirow{5}{*}{1024} & FP16 & - & 0 & 1 & 0.99  &26.53 & 0 & 1 & 1.05 & 26.33 \\
    \cmidrule{3-12}
    & & \cellcolor{blue!10}W8A8 & \cellcolor{blue!10}Ours & \cellcolor{blue!10}8.53 & \cellcolor{blue!10}0.67 & \cellcolor{blue!10}0.96  & \cellcolor{blue!10}26.54 & \cellcolor{blue!10}9.07 & \cellcolor{blue!10}0.64 & \cellcolor{blue!10}1.04 & \cellcolor{blue!10}26.07 \\
    \cmidrule{3-12}
    & & W4A16 & NF4(g=128) & 32.12 & 0.56 & 0.61  & 25.78 & 33.61 & 0.52 & 0.79 & 25.89 \\
    & & W4A16 & NF4(g=64) & 12.97 & 0.58 & 0.86  & 26.23 & 15.03 & 0.55 & 0.97 & 26.14 \\
    & & \cellcolor{blue!10}W4A8 & \cellcolor{blue!10}Ours & \cellcolor{blue!10}\textbf{9.64} & \cellcolor{blue!10}\textbf{0.62} & \cellcolor{blue!10}\textbf{1.00} &  \cellcolor{blue!10}\textbf{26.66} & \cellcolor{blue!10}\textbf{12.17}& \cellcolor{blue!10}\textbf{0.56} & \cellcolor{blue!10}\textbf{1.06} & \cellcolor{blue!10}\textbf{26.39}\\
    
    \midrule
    
    \end{tabular}
    \label{main_results}
\end{table*} 

\begin{figure*}[t]
    \centering
    \includegraphics[width=0.95\linewidth]{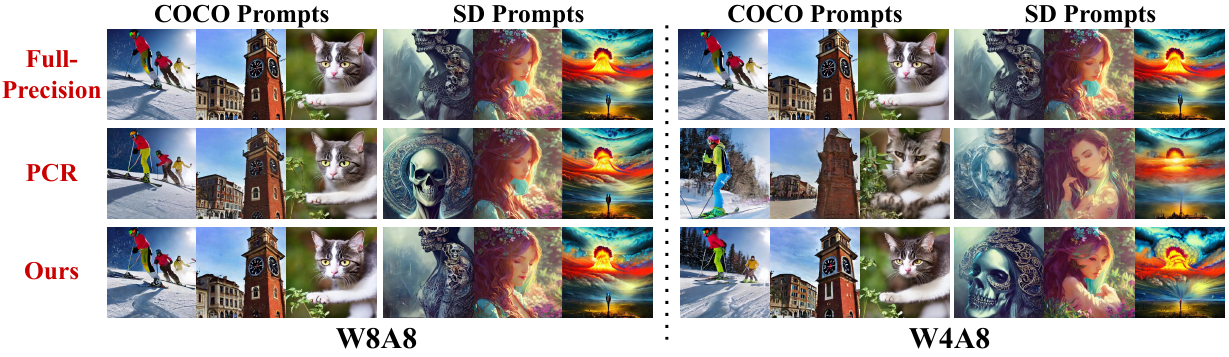}
    \caption{Stable Diffusion v1.4 512$\times$512 image generation using COCO prompts and Stable-Diffusion-Prompts.}
    \label{sd14_coco}
\end{figure*}

\begin{figure*}[t]
    \centering
    \includegraphics[width=0.95\linewidth]{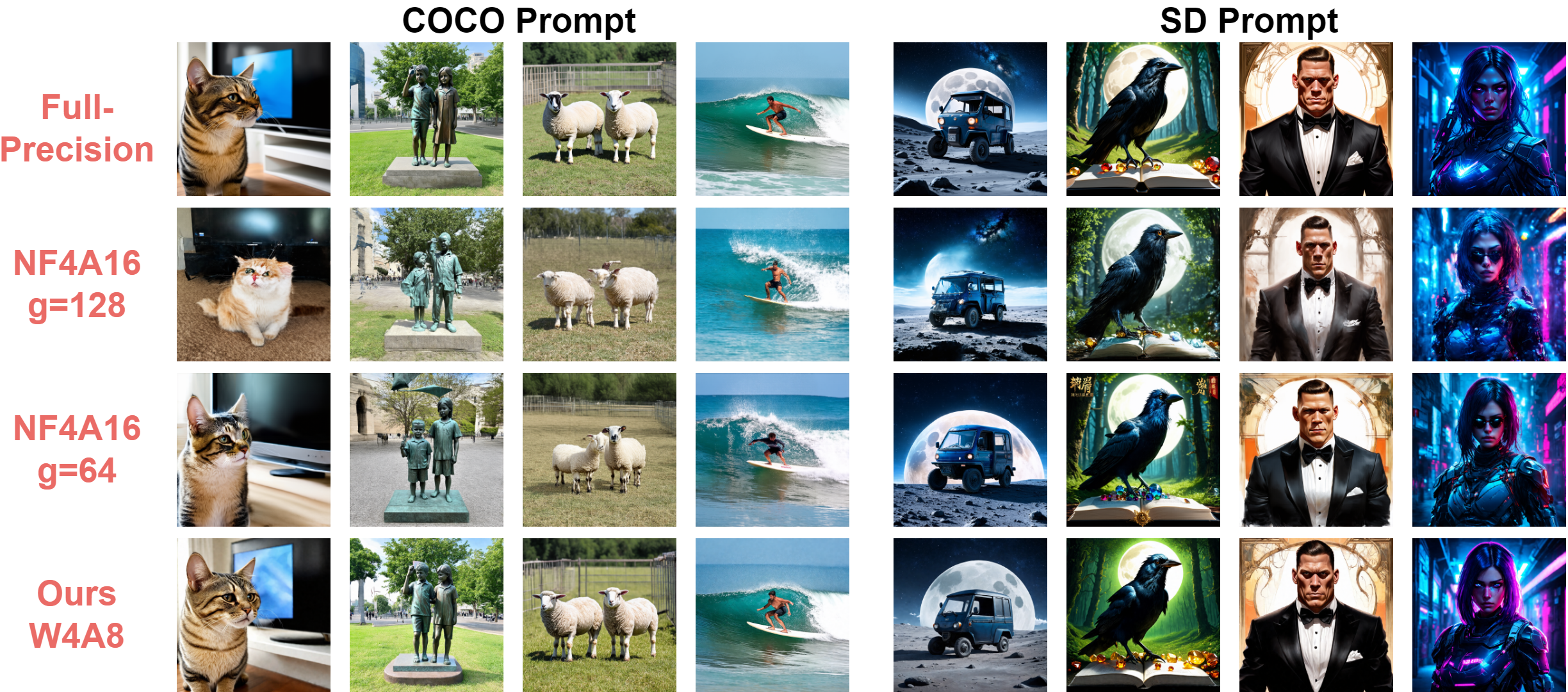}
    \caption{Stable Diffusion v3 1024$\times$1024 image generation using COCO prompts and Stable-Diffusion-Prompts.}
    \label{sd3}
\end{figure*}
\subsection{Objective Function}

We optimize quantized DDB $\epsilon_{q}$ to mimic the output of the floating-point DDB $\epsilon_{fp}$. Given the latent $\{x_{t1}\ldots x_{tn}\}$ at timestep $\{t_{1}\ldots t_{n}\}$, text embedding $\{p_{1}\ldots p_{n}\}$ from frozen text encoder, and precalculated projection $ep$, the output loss is defined as the mean squared error between the quantized and floating-point DDB outputs:
\begin{equation}
\mathcal{L}_{\text{out}} = \mathbb{E} \left[ \left\| \epsilon_{fp}(x_{t_{1}\ldots t_{n}}, p_{1\ldots n}, e_p) - \epsilon_{q}(x_{t_{1}\ldots t_{n}}, p_{1\ldots n}, e_p) \right\|_2^2 \right]
\end{equation}
where $\epsilon_{fp}$ and $\epsilon_{q}$ indicate the floating-point DDB and the quantized DDB, respectively.

To perform inter-layer distillation, the loss function of the feature maps by the sensitive layers is added:
\begin{equation}
\mathcal{L}_{\text{sen}} = \mathbb{E}\left[\left\| f_{fp}^s(x_{t_{1}\ldots t_{n}}, p_{1\ldots n}, e_p) - f_{q}^s(x_{t_{1}\ldots t_{n}}, p_{1\ldots n}, e_p) \right\|_2^2 \right]
\end{equation}
where $f_{fp}^s$ and $f_{q}^s$ indicate the floating-point and quantized feature maps of the sensitive layer, respectively.

The final loss function is: $\mathcal{L} = \mathcal{L}_{\text{out}} + \mathcal{L}_{\text{sen}}$ 
\section{Experiments}

\subsection{Experimental Setup}
\textbf{Datasets.} In this paper, we conduct experiments using two distinct datasets: COCO~\cite{lin2014microsoft} and Stable-Diffusion-Prompts. We utilize prompts from the COCO2017 training dataset to construct the latent dataset. In terms of evaluation, the process is twofold. Following~\cite{tang2024posttrainingquantizationtexttoimagediffusion}, firstly 5,000 prompts are selected from the COCO2017 validation dataset, which has been extensively employed in previous studies. Secondly, an additional 5,000 prompts from the Stable-Diffusion-Prompts dataset are used to assess the generalization capabilities of our quantized model in different prompt scenarios.

\textbf{Metrics.} 
We evaluate the generative results of quantized models through three dimensions: (1) Similarity, (2) Quality, and (3) Text-image alignment. We compute the Fréchet Inception Distance between images generated by the quantized model and floating-point model (FID-to-FP~\cite{tang2024posttrainingquantizationtexttoimagediffusion}) to compute distributional similarity,  alongside Structural Similarity (SSIM~\cite{Wang2004ssim}) for pixel-level visual correspondence. Image quality is assessed using Image Reward (IR~\cite{xu2023imagereward}), a human-preference-aligned metric that correlates with subjective ratings. Additionally, we use CLIP score~\cite{hessel2021clipscore} to evaluate the matching degree between images and prompts. The evaluation code is adopted from ~\cite{teng2023relaydiffusionunifyingdiffusion} and ~\cite{pyiqa}.

\textbf{Baselines and implementation.} We compare our proposed approach against advanced techniques: Q-diffusion~\cite{li2023q}, PTQ4DM~\cite{shang2023post},  PCR~\cite{tang2024posttrainingquantizationtexttoimagediffusion}, and NF4~\cite{dettmers2023qlora}. Results are obtained from PCR~\cite{tang2024posttrainingquantizationtexttoimagediffusion} or reproduced. We employ the Stable Diffusion v1-4, Stable Diffusion v2-1, Stable Diffusion XL 1.0, and Stable Diffusion v3, both sourced from Hugging Face. We compare extensively with PCR~\cite{tang2024posttrainingquantizationtexttoimagediffusion} and align the quantization and generation settings with it. We use per-channel weight quantization and static per-tensor activation quantization. Except for special declaration, we adopt the default noise scheduler and Classifier-Free Guidance (CFG) scale. All experiments are conducted using a single NVIDIA A100.

\subsection{Main Results}
\textbf{Quantitative results} We demonstrate metric evaluation results in Tab.~\ref{main_results} across various models and precision levels. 

on Stable Diffusion v1.4 \& v2.1, Our method consistently outperforms PCR under the W4A8 configuration across all metrics, without employing the activation relaxation strategy in PCR (PCR relaxes 20\% timesteps' activation bitwidth to 10-bit, so it's actually 8.4bit). Our approach achieves significantly higher similarity to FP16 baselines with negligible degradation in IR scores. Under W8A8, while our FID-FP is marginally higher than PCR’s W8A8.4, our method dominates other metrics.

For SDXL, we validate PCR’s observation that activation bitwidth relaxation is critical for text-image alignment (CLIP Score). Adopting PCR’s settings, our method still delivers substantial improvements over PCR, with empirical evidence showing that CLIP Score degradation correlates strongly with IR drop.

Additionally, we validate our method on MM-DiT based Stable Diffusion v3. Since previous methods haven't tried on sd3, we establish NF4 (4-bit non-uniform weight quantization) as the baseline. Our W4A8 surpasses NF4’s W4A16 at both 512×512 and 1024×1024 resolutions, demonstrating scalability to advanced architectures.

\textbf{Visual results.} We show some corresponding qualitative comparisons on SD1.4 and SD3 in Fig.~\ref{sd14_coco} and Fig.~\ref{sd3}. Previous methods, when quantized to 4-bit, result in noticeable style changes in the generated images compared to those produced by the floating-point model. Such changes include but are not limited to, alterations in scene layout and facial features, loss of color and object, and the blending of multiple objects. In contrast, the images generated by our method are consistently of high fidelity.

\subsection{Dataset Generation Analysis}
In this section, We will discuss the trade-off of some crucial hyperparameters in the latent dataset generation. 
\begin{table}[t]
    \centering
    \renewcommand{\arraystretch}{0.9}
    \caption{Comparison of different sampling strategies for latent dataset generation.}
    \begin{tabular}{ccccc}
    \toprule
    Methods & Prompts & Size & Time & FID-to-FP\\
    \midrule
    50steps/prompt & 4000 & 6G & 1.8h & 9.44 \\
    \midrule
    1steps/prompt & 20000 & 0.6G & 4h & 9.46 \\
    \bottomrule
    \end{tabular}
    \label{sample_compare}
\end{table}
\begin{figure}
    \centering
    \includegraphics[width=0.9\linewidth]{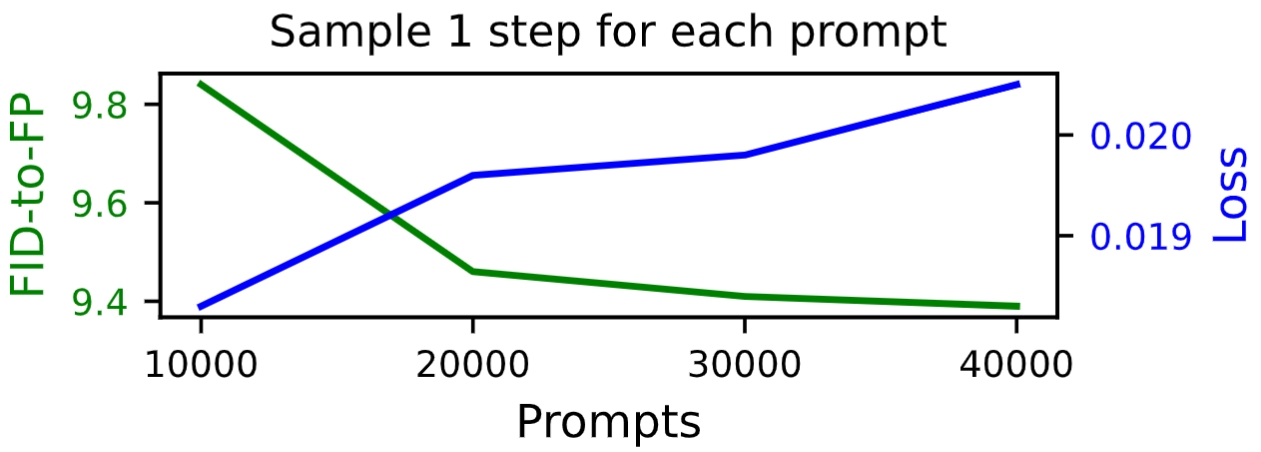}
    \caption{Comparison of Loss and FID-to-FP Curves under different number of prompts. The overfitting problem is obvious.}
    \label{loss_fid}
\end{figure}
\begin{table}[t]
    \centering
    \caption{Efficiency comparison on Stable Diffusion v1-4 and Stable Diffusion XL. (a) denotes sampling 50 steps per prompt, and (b) denotes sampling 1 step per prompt.}
    \small
    \renewcommand{\arraystretch}{0.9}
    \begin{tabular}{ccc|c|c|c}
    \toprule
    \multicolumn{2}{c}{\multirow{2}{*}{Methods}} & \multicolumn{4}{c}{Time Cost}\\
    \cmidrule{3-6}
    & & \multicolumn{2}{c|}{SD1.4} & \multicolumn{2}{c}{SDXL}\\
    \midrule
    \multicolumn{2}{c}{PCR} & \multicolumn{2}{c|}{$\approx$13h} & \multicolumn{2}{c}{$\approx$25h} \\
    \midrule
    \multirow{3}{*}{Ours} & \multirow{2}{*}{\makecell{Gen Dataset\\(Once for all)}} & (a) & (b) & (a) & (b) \\
    \cmidrule{3-6}
    & & 1.8h &$\approx$4h & 2.5h & $\approx$5.5h\\
    \cmidrule{2-6}
    & Train & \multicolumn{2}{c|}{$\approx$4.5h} & \multicolumn{2}{c}{$\approx$7.5h}\\
    \bottomrule
          
    \end{tabular}
    \label{efficiency}
\end{table}

\textbf{More prompts or more timesteps?} In the dataset generation process, we can randomly sample varying amounts of latent for each prompt. For comparison, two datasets have been constructed. The first dataset comprises 4000 prompts, with 50 latent per prompt. The second consists of 20000 prompts, each prompt with just 1 latent. As demonstrated in Tab.~\ref{sample_compare}, despite the first dataset having 10$\times$ more latent, it exhibits a similar FID-to-FP. We can infer that a dataset with more prompts is more resistant to overfitting and has a smaller size but with a longer generation time. Given the comparable outcomes of the two strategies, we choose the latter dataset for smaller storage

\textbf{Length of dataset.} Given our limited training iterations, prompt quantity plays a critical role. Figure~\ref{loss_fid} clearly demonstrates the overfitting challenges with smaller datasets. To mitigate this, we employ default settings of 20,000 prompts for Stable Diffusion v1.4 and 10,000 prompts for Stable Diffusion XL, using 1-step sampling. As summarized in Table~\ref{efficiency}, our method achieves substantial training efficiency gains over PCR, particularly for larger architectures like Stable Diffusion XL.

\subsection{Ablation Study}
Tab.~\ref{ablation_study} presents our ablation study on Stable Diffusion v2-1 with W4A8 quantization. The 'Base' configuration represents the original serial pipeline as illustrated in Fig.~\ref{pipeline_compare}. We systematically evaluate each component's contribution by gradually adding each of them.
The results clearly show that each proposed component contributes meaningfully to the improvement of fidelity. Notably, the Serial-to-Parallel pipeline exhibits the most pronounced effect. Our method incorporates all these components effectively.
\begin{table}[t]
\centering
\caption{Ablation results on the COCO prompts for Stable Diffusion v2-1 under W4A8 settings. 'S2P' denotes the Serial-to-Parallel pipeline, 'Time' denotes time information precalculation, 'Mstep' denotes multi-step activation quantization, 'Distill' denotes inter-layer distillation, and 'Freeze' denotes selective freezing.}
\renewcommand{\arraystretch}{0.9}
\begin{tabular}{lccccc}
\toprule
\renewcommand{\arraystretch}{0.9}
Method & FID-FP$\downarrow$ &LPIPS$\downarrow$ & SSIM$\uparrow$ &IR$\uparrow$ & Clip$\uparrow$ \\ 
\midrule
FP32 & 0.00 &0.00 & 1 & 0.32 & 26.07\\
\midrule
Base & 13.68 & 0.46 & 0.50 & 0.09 & 25.90 \\
+S2P & 12.04 & 0.40 & 0.55 & 0.27 & 25.92 \\
+Time & 11.95 & 0.40 & 0.56 & 0.27& 26.01 \\
+Mstep & 11.90& 0.39 & 0.56 & 0.28& 25.99 \\
+Distill & 11.71& 0.35& 0.58 & 0.26& 26.04 \\
+Freeze & \textbf{11.60} & \textbf{0.33} & \textbf{0.60} & \textbf{0.33} & \textbf{26.08} \\
\bottomrule
\end{tabular}
\label{ablation_study}
\end{table}

\section{Conclusion}
This research explores the application of quantization to Stable Diffusion models. In this paper, we propose an efficient quantization framework for Stable Diffusion models aiming for higher generation consistency. We introduce a Serial-to-Parallel pipeline which not only considers the consistency of the training process and the inference process but also ensures the training stability. With the aid of multi-timestep activation quantization, time information precalculation, inter-layer distillation, and selective freezing strategies, high-fidelity generation is guaranteed. Extensive experiments demonstrate that our method generates high-fidelity and high-quality figures within a shorter time and outperforms state-of-the-art techniques. These advancements highlight the framework's strong potential for enabling efficient edge deployment of high-quality diffusion models.

\bibliographystyle{ACM-Reference-Format}
\bibliography{sample-base}

\end{document}